\documentclass[sigconf]{acmart}

\usepackage{subfigure}
\usepackage{algorithm}
\usepackage{algorithmic}
\usepackage{xcolor}
\usepackage{enumitem}
\usepackage{amsmath}
\usepackage{bbm}

\newcommand{{\method}}{\sf GraphSMOTE}
\DeclareMathOperator*{\argmin}{argmin}
\DeclareMathOperator*{\argmax}{argmax}

\AtBeginDocument{%
  \providecommand\BibTeX{{%
    \normalfont B\kern-0.5em{\scshape i\kern-0.25em b}\kern-0.8em\TeX}}}

\copyrightyear{2021}
\acmYear{2021}
\setcopyright{acmcopyright}\acmConference[WSDM '21]{Proceedings of the Fourteenth ACM International Conference on Web Search and Data Mining}{March 8--12, 2021}{Virtual Event, Israel}
\acmBooktitle{Proceedings of the Fourteenth ACM International Conference on Web Search and Data Mining (WSDM '21), March 8--12, 2021, Virtual Event, Israel}
\acmPrice{15.00}
\acmDOI{10.1145/3437963.3441720}
\acmISBN{978-1-4503-8297-7/21/03}

\begin{document}

\fancyhead{}
\title{GraphSMOTE: Imbalanced Node Classification on Graphs \\ with Graph Neural Networks}




\author{Tianxiang Zhao, Xiang Zhang, Suhang Wang}
\email{{tkz5084, xzz89, szw494}@psu.edu}
\affiliation{%
   \institution{College of Information Science and Technology, Penn State University}
   \city{State College}
   \country{The USA}
}


\begin{abstract}
Node classification is an important research topic in graph learning. Graph neural networks (GNNs) have achieved state-of-the-art performance of node classification. However, existing GNNs address the problem where node samples for different classes are balanced; while for many real-world scenarios, some classes may have much fewer instances than others. Directly training a GNN classifier in this case would under-represent samples from those minority classes and result in sub-optimal performance. Therefore, it is very important to develop GNNs for imbalanced node classification. However, the work on this is rather limited. Hence, we seek to extend previous imbalanced learning techniques for i.i.d data to the imbalanced node classification task to facilitate GNN classifiers. In particular, we choose to adopt synthetic minority over-sampling algorithms, as they are found to be the most effective and stable. This task is non-trivial, as previous synthetic minority over-sampling algorithms fail to provide relation information for newly synthesized samples, which is vital for learning on graphs. Moreover, node attributes are high-dimensional. Directly over-sampling in the original input domain could generates out-of-domain samples, which may impair the accuracy of the classifier. We propose a novel framework, {\method}, in which an embedding space is constructed to encode the similarity among the nodes. New samples are synthesize in this space to assure genuineness. In addition, an edge generator is trained simultaneously to model the relation information, and provide it for those new samples. This framework is general and can be easily extended into different variations. The proposed framework is evaluated using three different datasets, and it outperforms all baselines with a large margin. 

\end{abstract}

\maketitle

\section{Introduction}

Recent years have witnessed great improvements in learning from graphs with the developments of graph neural networks(GNNs)~\cite{Kipf2017SemiSupervisedCW,Hamilton2017InductiveRL,xu2018powerful}. One typical task is semi-supervised node classification~\cite{yang2016revisiting}, where we have a large graph with a small ratio of nodes labeled. A classifier can be trained on those supervised nodes, and be used to classify other nodes during testing. GNNs have obtained state-of-the-art performance in this task, and is developing rapidly. For example, GCN~\cite{Kipf2017SemiSupervisedCW} exploits features in the spectral domain efficiently by using a simplified first-order approximation; GraphSage~\cite{Hamilton2017InductiveRL} utilizes features in the spatial domain and is better at adapting to diverse graph topology. 
Despite all these progresses, existing work mainly focus on the setting that node classes are balanced.

\begin{figure}[t!]
  \centering
  \subfigure[Bot detection task]{
		\label{fig:example-bot}
		\includegraphics[width=0.22\textwidth]{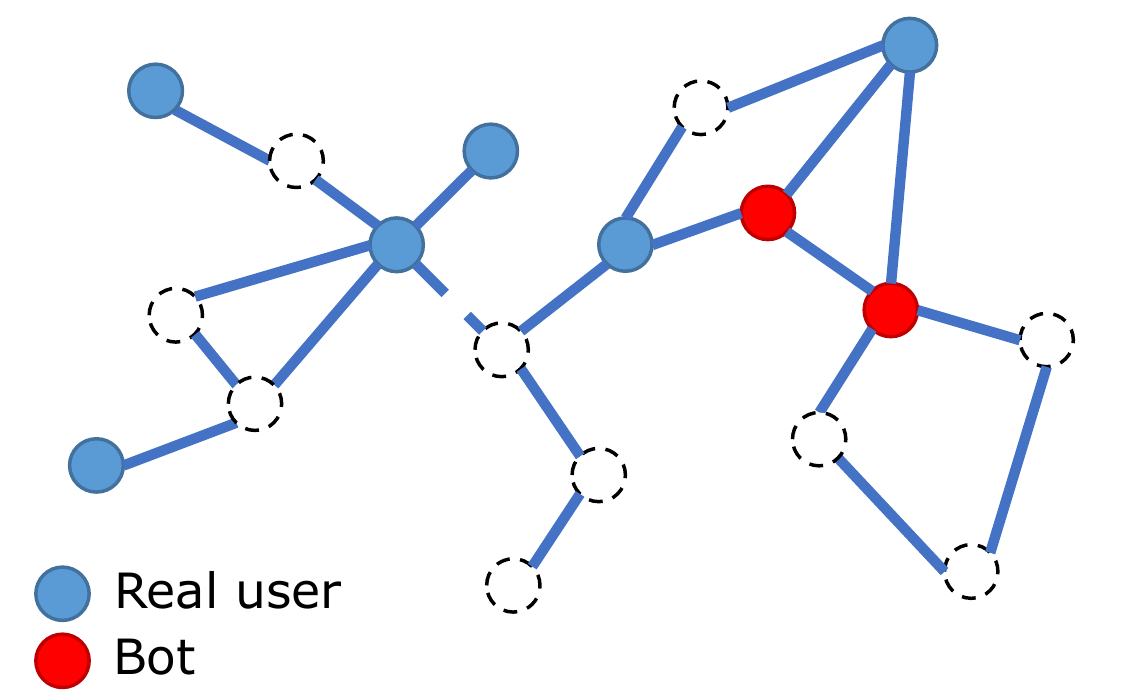}}
    \subfigure[After over-sampling]{
		\label{fig:example-after}
		\includegraphics[width=0.22\textwidth]{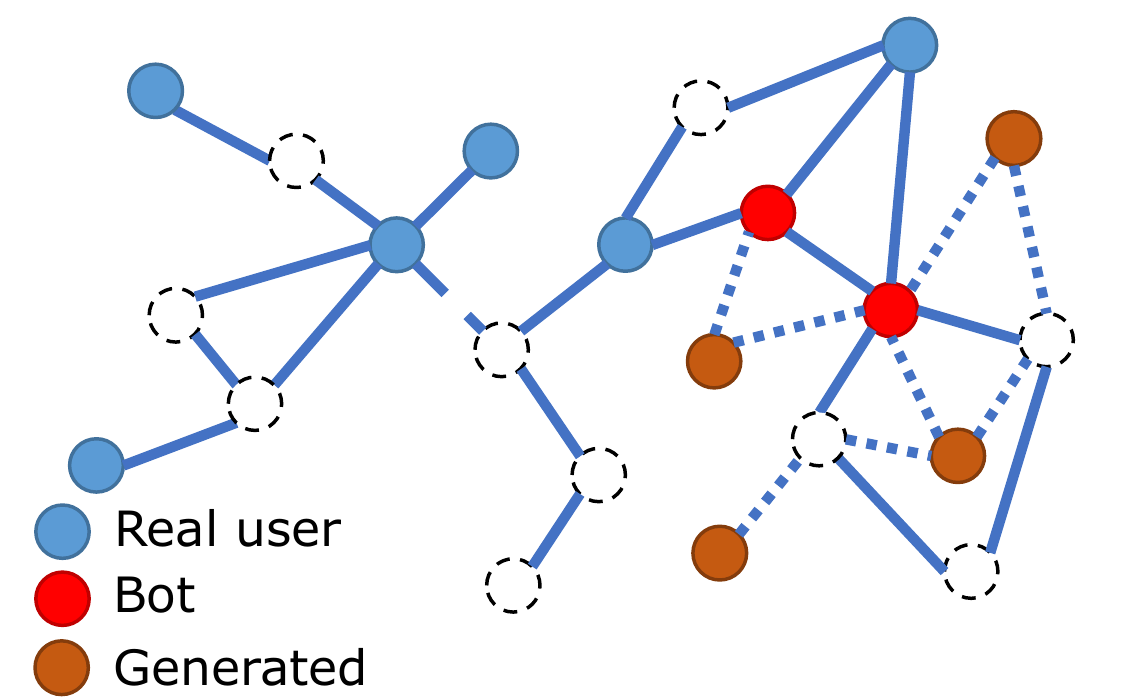}}
		
    \vskip -1em
    \caption{An example of bot detection on a social network, and the idea of over-sampling. Note that the over-sampling is in the latent space.} \label{fig:example}
  \setlength{\abovecaptionskip}{0cm}
\end{figure}

In many real-world applications, node classes could be imbalanced in graphs, i.e., some classes have significantly fewer samples for training than other classes. For example, for fake account detection~\cite{mohammadrezaei2018identifying,zhao2009botgraph}, the majority of users in a social network platform are benign users while only a small portion of them are bots. Similarly, topic classification for website pages~\cite{wang2020network} could also suffer from this problem, as the materials for some topics are scarce, comparing to those on-trend topics. 
Thus, we are often faced with imbalanced node classification problem. An example of the imbalanced node classification problem is shown in Figure~\ref{fig:example-bot}.  Each blue node refers to a real user, each red node refers to a fake user, and the edges denote the friendship. The task is to predict whether those unlabeled users in dashes are real or fake. The classes are imbalanced in nature, as fake users are often less than $1\%$ of all the users. The semi-supervised setting further magnifies the class imbalanced issue as we are only given limited labeled data, which makes the number of labeled minority samples extremely small.

The imbalanced node classification brings challenges to existing GNNs because the majority classes could dominate the loss function of GNNs, which makes the trained GNNs over-classify those majority classes and become unable to predict accurately for samples from minority classes. This issue impedes the adoption of GNNs for many real-world applications with imbalanced class distribution such as malicious account detection. Therefore, it is important to develop GNNs for class imbalanced node classification.

In machine learning domain, traditional class imbalance problem has been extensively studied. Algorithms can be summarized into three groups: data-level approaches, algorithm-level approaches, and hybrid approaches. Data-level approaches seek to make the class distribution more balanced, using over-sampling or down-sampling techniques\cite{more2016survey,chawla2002smote}; algorithm-level approaches typically introduce different mis-classification penalties or prior probabilities for different classes~\cite{elkan2001foundations,ling2008cost,zhou2005training}; and hybrid approaches~\cite{chawla2003smoteboost,liu2008exploratory} combine both of them. However, directly applying them to graphs may get sub-optimal results. Relation is the key information needed to be exploited in graph data, and under-representation of minority samples would impair not only their embedding quality, but also the knowledge exchange processes across neighboring nodes. Previous algorithms fail to address that due to their i.i.d assumption, taking each sample as independent.

Therefore, in this work, we study a novel problem of exploring synthetic minority oversampling for imbalanced node classification with GNNs~\footnote{Code available at https://github.com/TianxiangZhao/GraphSmote}. The idea is shown in Figure~\ref{fig:example-after}. Previous algorithms are not readily applicable to graphs, due to two-folded reasons. First, it is difficult to generate relation information for synthesized new samples. Mainstream oversampling techniques~\cite{more2016survey} use interpolation between target example and its nearest neighbor to generate new training examples. However, interpolation is improper for edges, as they are usually discreet and sparse. Interpolation could break down the topology structure. Second, synthesized new samples could be of low quality. Node attributes are high-dimensional, and directly interpolating on them would easily generate out-of-domain examples, which are not beneficial for training the classifier. 

Targeting at these two problems, we extend previous over-sampling algorithms to a new framework, {\method}, in order to cope with graphs. The modifications are mainly at two places. First, we propose to obtain new edges between generated samples and existing samples with an edge predictor. This predictor can learn the genuine distribution of edges, and hence can be used to produce reliable relation information among samples. Second, we propose to perform interpolation at the intermediate embedding space of a GNN network, inspired by ~\cite{ando2017deep}. In this intermediate embedding space, the dimensionality is much lower, and the distribution of samples from the same class would be more dense. As intra-class similarity as well as inter-class differences would have been captured by previous layers, interpolation can be better trusted to generate in-domain samples. Concretely, we propose a new framework in which graph auto-encoding task and node classification task are combined together. These two tasks share the same feature extractor, and oversampling is performed at the output space of that module, as shown in Figure~\ref{fig:model_architecture}. The main contributions are:
\begin{itemize}
    \item We propose to study a novel problem, node class imbalance problem for learning on graphs. It has many real-world applications, and this paper is the first work focusing on this task as far as we know.
    \item We design a new framework which extends previous over-sampling algorithms to work for graph data. It addresses the deficiencies of previous methods, by generating more natural nodes as well as relation information. Besides, it is general and easy to extend.
    \item Experiments are performed on three datasets, and {\method} outperforms all baselines with a large gap. Extensive analysis of our model's behavior as well as recommended settings are also presented.
\end{itemize}

The rest of the paper are organized as follows. In Sec.~\ref{sec:related_work}, we review related work. In Sec.~\ref{sec:problem_definition}, we formally define the problem. In Sec.~\ref{sec:methodology}, we give the details of {\method}. In Sec.~\ref{sec:experiments}, we conduct experiments to evaluate the effectiveness of {\method}. In Sec.~\ref{sec:conclusion}, we conclude with future work.

\section{Related Work} \label{sec:related_work}
In this section, we briefly review related works, which include graph neural networks and class imbalance problem.

\subsection{Class Imbalance Problem}
Class imbalance is common in real-world applications, and has long been a classical research direction in the machine learning domain. Plenty of tasks suffer from this problem, like medical diagnosis~\cite{mac2002problem,grzymala2004approach} or fine-grained image classification~\cite{peng2017object,van2017inaturalist}. Classes with larger number of instances are usually called as majority classes, and those with fewer instances are usually called as minority classes. The countermeasures against this problem can generally be classified into three groups, i.e., algorithm-level, data-level and hybrid. 

The first group of methods are data-level, seeking to directly adjust class sizes through over- or under-sampling. The vanilla form of over-sampling is replicating existing samples. It reduces this imbalance, but can lead to over-fitting as no extra information is introduced. SMOTE~\cite{chawla2002smote} addresses this problem by generating new samples, performing interpolation between samples in minority classes and their nearest neighbors. SMOTE is the most popular over-sampling approach, and many extensions are proposed on top of it to make the interpolation process more effective. For example, Borderline-SMOTE~\cite{han2005borderline} limits over-sampling to samples near the borderline of classes, which are believed to be more informative. Safe-Level-SMOTE~\cite{bunkhumpornpat2009safe} computes the safe level for each interpolation direction using majority class neighbors, in order to make the generated new samples safer. Cluster-based Over-sampling~\cite{jo2004class} first clusters samples into different groups, than over-samples each group separately, considering that small districts often exist in the input space. Besides, ~\cite{ando2017deep} extends over-sampling to work with CNNs, through interpolation in an embedding space. Under-sampling discards some samples from majority classes, which can also make classes balanced, but at the price of losing some information. To overcome this deficiency, many extensions are proposed to remove only redundant samples, like ~\cite{kubat1997addressing,barandela2004imbalanced}. The second group of methods are algorithm-level. Cost sensitive learning~\cite{zhou2005training,ling2008cost} generally constructs a cost matrix to assign different mis-classification penalties for different classes. Its effect is similar to vanilla over-sampling. ~\cite{parambath2014optimizing} proposes an approximation to F measurement, which can be directly optimized by gradient propagation.  Threshold moving~\cite{lawrence1998neural} modifies the inference process after the classifier is trained, by introducing a prior probability for each class. Through these approaches, the importance of minority classes can be increased. The last group are hybrid approaches, which combine multiple algorithms from one or both aforementioned categories. ~\cite{liu2008exploratory} uses a group of classifiers, each one is trained on a subset of majority classes and minority classes. ~\cite{chawla2003smoteboost} combines boosting with SMOTE approach, and ~\cite{he2009learning} combines over-sampling with cost sensitive learning. ~\cite{sun2007cost} introduces three cost-sensitive boosting approaches, which iteratively updates the impact of each class in together with the AdaBoost parameters.

Some systematic analysis of them have found that synthetic minority oversampling techniques such as SMOTE are the most popular and effective approaches for addressing  class imbalance~\cite{buda2018systematic,johnson2019survey}. However, existing work are overwhelmingly dedicated to i.i.d data. They cannot be directly applied to graph structured data because: (i) the synthetic node generation on the raw feature space cannot take the graph information into consideration; and (ii) the generated nodes doesn't have links with the graph, which cannot facilitate the graph based classifier such as GNNs. Hence, in this work, we focus on extending SMOTE into graph domain for GNNs.

\subsection{Graph Neural Network}
In recent years, with the increasing requirements of learning on non-Euclidean space and modeling rich relation information among samples, graph neural networks (GNNs) have received much more attention and are developing rapidly. GNNs generalize convolutional neural networks to graph structured data and have shown great ability in modeling graph structured data. Current GNNs follow a message-passing framework, which is composed of pattern extraction and interaction modeling within each layer~\cite{gilmer2017neural}. Generally, 
existing GNN frameworks can be categorized into two categorizes, i.e., spectral-based~\cite{bruna2013spectral,Tang2019ChebNetEA,Kipf2017SemiSupervisedCW,Hamilton2017InductiveRL} and spatial-based~\cite{duvenaud2015convolutional,atwood2016diffusion}. 

Spectral-based GNNs defines the convolution operation in the Fourier domain by computing the eigendecomposition of the graph Laplacian. Early work~\cite{bruna2013spectral} in this domain involves extensive computation, and is time-consuming. To accelerate, ~\cite{Tang2019ChebNetEA} adopts Chebyshev Polynomials to approximate spectral kernels, and enforces locality constraints by truncating only top-k terms. GCN~\cite{Kipf2017SemiSupervisedCW} takes a further step by preserving only top-2 terms, and obtains a more simplified form. GCN is one of the most widely-used GNN currently. However, all spectral-based GNNs suffer from the generalization problem, as they are dependent on the Laplacian eigenbasis~\cite{Zhou2018GraphNN}. Hence, they are usually applied in the transductive setting, training and testing on the same graph structure. Spatial-based GNNs are more flexible and have stronger in generalization ability. They implement convolutions basing on the neighborhoods of each node. As each node could have different number of neighbors, Duvenaud et al.,~\cite{duvenaud2015convolutional} uses multiple weight matrices, one for each degree. ~\cite{atwood2016diffusion} proposes a diffusion convolution neural network, and ~\cite{niepert2016learning} adopts a fixed number of neighbors for each sample. A more popular model is GraphSage~\cite{Hamilton2017InductiveRL}, which samples and aggregates embedding from local neighbors of each sample. More recently, ~\cite{xu2018powerful} extends expressive power of GNNs to that of WL test, and ~\cite{You2019PositionawareGN} introduce a new GNN layer that can encode node positions.

Despite the success of various GNNs, existing work doesn't consider the class imbalance problem, which widely exists in real-world applications and could significantly reduce the performance of GNNs. Thus, we study a novel problem of synthetic minority oversampling on graphs to facilitate the adoption of GNNs for class imbalance node classification.

\section{Problem Definition} \label{sec:problem_definition}
In this work, we focus on semi-supervised node classification task on graphs, in the transductive setting. As shown in Figure~\ref{fig:example}, we have a large network of entities, with some labeled for training. Both training and testing are performed on this same graph. Each entity belongs to one class, and the distribution of class sizes are imbalanced. This problem has many practical applications. For example, the homophily in social networks which results in the under-representation of minority groups, malicious behavior or fake user accounts on social networks which are outnumbered by normal ones, and linked web pages in knowledge base where materials for some topics are limited.

Throughout this paper, we use $\mathcal{G} = \{\mathcal{V}, \mathbf{A}, \mathbf{F}\}$ to denote an attributed network, where $\mathcal{V}=\{v_1,\dots,v_{n}\}$ is a set of $n$ nodes. $\mathbf{A} \in \mathbb{R}^{n \times n}$ is the adjacency matrix of $\mathcal{G}$, and $\mathbf{F} \in \mathbb{R}^{n \times d}$ denotes the node attribute matrix, where $\mathbf{F}[j,:] \in \mathbb{R}^{1 \times d}$ is the node attributes of node j and $d$ is the dimension of the node attributes. $\mathbf{Y} \in \mathbb{R}^{n}$ is the class information for nodes in $\mathcal{G}$. During training, only a subset of $\mathbf{Y}$, $\mathbf{Y}_L$, is available, containing the labels for node subset $\mathcal{V}_L$. There are $m$ classes in total, $\{\mathcal{C}_1, \dots, \mathcal{C}_m\}$. $|\mathcal{C}_i|$ is the size of $i$-th class, referring to the number of samples belong to that class. We use imbalance ratio, $\frac{min_{i}(|\mathcal{C}_i|)}{max_{i}(|\mathcal{C}_i|)}$, to measure the extent of class imbalance. In the imbalanced setting, imbalance ratio of $\mathbf{Y}_L$ is small.

\vspace{0.5em}
\noindent{}\textit{Given $\mathcal{G}$ whose node class set is imbalanced, and labels for a subset of nodes $\mathcal{V}_{L}$, we aim to learn a node classifier $f$ that can work well for both majority and minority classes, i.e.,
\begin{equation}
    f(\mathcal{V}, \mathbf{A}, \mathbf{F}) \rightarrow \mathbf{Y}
\end{equation}
}

\section{Methodology} \label{sec:methodology}
\begin{figure}[t!]
  \centering
    \includegraphics[width=0.49\textwidth]{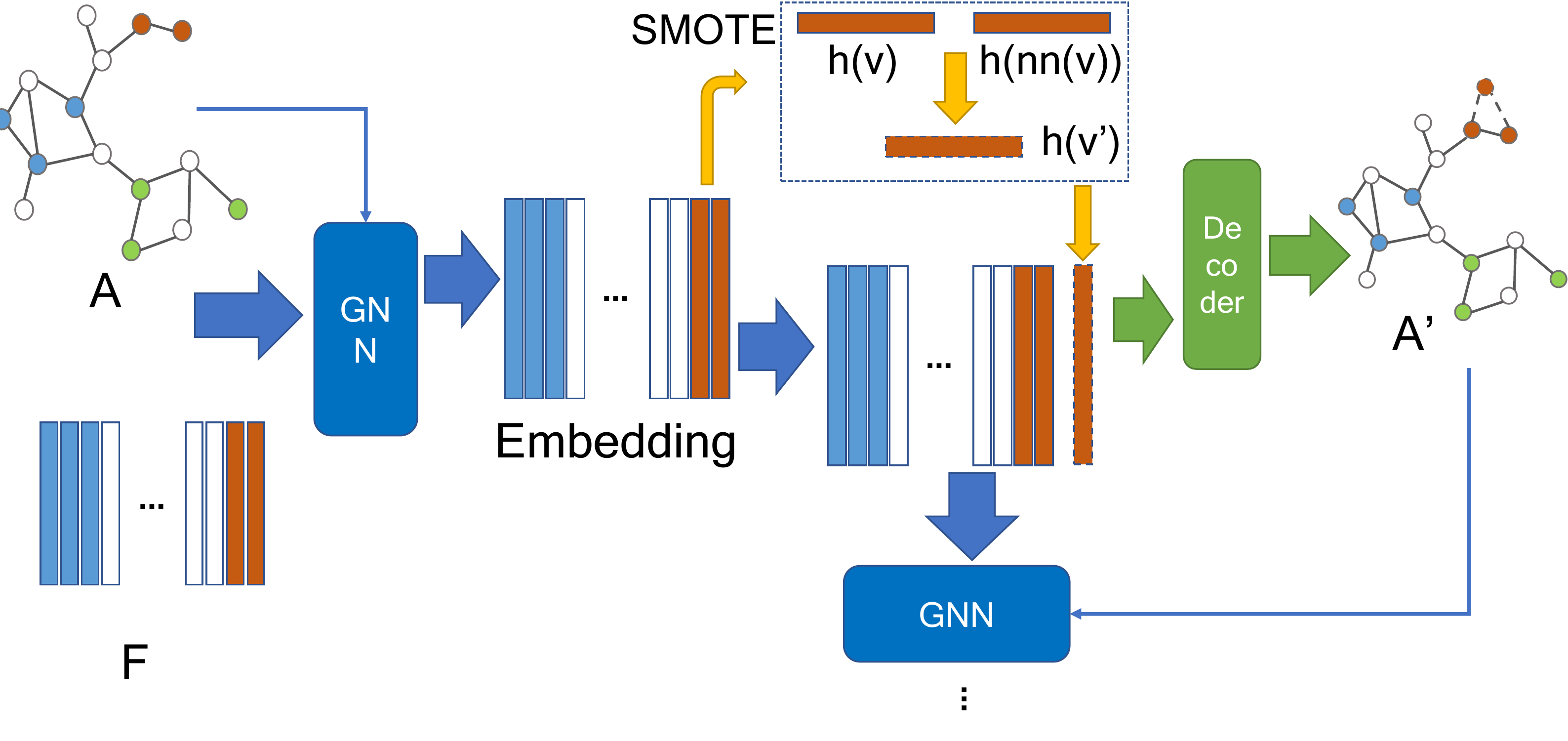}
    \vskip -1em
    \caption{Overview of the framework} \label{fig:model_architecture}
  \setlength{\abovecaptionskip}{0cm}
\end{figure}

In this section, we give the details of the proposed framework
{\method}. The main idea of {\method} is to generate synthetic minority nodes through interpolation in an expressive embedding space acquired by the GNN-based feature extractor, and use an edge generator to predict the links for the synthetic nodes, which forms an augmented balanced graph to facilitate node classification by GNNs. An illustration of the proposed framework is shown in Figure~\ref{fig:model_architecture}. {\method} is composed of four components: (i) a GNN-based feature extractor (encoder) which learns node representation that preserves node attributes and graph topology to facilitate the synthetic node generation; (ii) a synthetic node generator which generates synthetic minority nodes in the latent space; (iii) an edge generator which generate links for the synthetic nodes to from an augmented graph with balanced classes; and (iv) a GNN-based classifier which performs node classification based on the augmented graph.  Next, we give the details of each component.

\subsection{Feature Extractor}

One way to generate synthetic minority nodes is to directly apply SMOTE on the raw node feature space. However, this will cause several problems: (i) the raw feature space could be sparse and high-dimensional, which makes it difficult to find two similar nodes of the same class for interpolation; and (ii) it doesn't consider the graph structure, which can result in sub-optimal synthetic nodes. Thus, instead of directly do synthetic minority over-sampling in the raw feature space, we introduce an feature extractor learn node representations that can simultaneously capture node properties and graph topology. Generally, the node representations should reflect inter-class and intra-class relations of samples. Similar samples should be closer to each other, and dissimilar samples should be more distant. In this way, when performing interpolation on minority node with its nearest neighbor, the obtained embedding would have a higher probability of representing a new sample belonging to the same minority class. In graphs, the similarity of nodes need to consider node attributes, node labels, as well as local graph structures. Hence, we implement it with GNN, and train it on two down-stream tasks, edge prediction and node classification. 

The feature extractor can be implemented using any kind of GNNs. In this work, we choose GraphSage as the backbone model structure because it is effective in learning from various types of local topology, and generalizes well to new structures. It has been observed that too deep GNNs often lead to sub-optimal performance, as a result of over-smoothing and over-fitting. Therefore, we adopt only one GraphSage block as the feature extractor.
Inside this block, the message passing and fusing process can be written as:
\begin{equation}
\mathbf{h}_{v}^{1}  = \sigma(\mathbf{W}^1 \cdot CONCAT(\mathbf{F}[v,:], \mathbf{F} \cdot \mathbf{A}[:,v])),
\end{equation}
$\mathbf{F}$ represents input node attribute matrix and $\mathbf{F}[v,:]$ represents attribute for node $v$. $\mathbf{A}[:,v]$ is the $v$-th column in adjacency matrix, and $\mathbf{h}_{v}^{1}$ is the obtained embedding for node $v$. $\mathbf{W}^1$ is the weight parameter, and $\sigma$ refers to the activation function such as ReLU. 

\subsection{Synthetic Node Generation}
After obtaining the representation of each node in the embedding space constructed by the feature extractor, now we can perform over-sampling on top of that. We seek to generate the expected representations for new samples from the minority classes. 
In this work, to perform over-sampling, we adopt the widely used SMOTE algorithm, which augments vanilla over-sampling via changing repetition to interpolation. We choose it due to its popularity, but our framework can also cope with other over-sampling approaches as well. The basic idea of SMOTE is to perform interpolation on samples from the target minority class with their nearest neighbors in the embedding space that belong to the same class. Let $\mathbf{h}_v^1$ be a labeled minority nodes with label as $Y_{v}$. The first step is to finds the closest labeled node of the same class as $\mathbf{h}_v^1$, i.e., 
\begin{equation}
    nn(v) = \argmin_{u} \|\mathbf{h}_{u}^{1} - \mathbf{h}_{v}^{1}\|, \quad \text{s.t.}  \quad \mathbf{Y}_u = \mathbf{Y}_v
\end{equation}
$nn(v)$ refers to the nearest neighbor of $v$ from the same class, measured using Euclidean distance in the embedding space. With the nearest neighbor, we can generate synthetic nodes as
\begin{equation}
\begin{aligned}
    \mathbf{h}_{v'}^{1} &= (1-\delta) \cdot \mathbf{h}_{v}^{1} + \delta \cdot \mathbf{h}_{nn(v)}^{1},
\end{aligned}
\end{equation}
where $\delta$ is a random variable, following uniform distribution in the range $[0,1]$. Since $\mathbf{h}_{v}^{1}$  and $\mathbf{h}_{nn(v)}^{1}$ belong to the same class and are very close to each other, the generated synthetic node $\mathbf{h}_{v'}^{1}$ should also belong to the same class. In this way, we can obtain labeled synthetic nodes.

For each minority class, we can apply SMOTE to generate syntetic nodes. We use a hyper-parameter, over-sampling scale, to control the amount of samples to be generated for each class. Through this generation process, we can make the distribution of class size more balanced, and hence make the trained classifier perform better on those initially under-represented classes.

\subsection{Edge Generator}

Now we have generated synthetic nodes to balance the class distribution. However, these nodes are isolated from the raw graph $\mathcal{G}$ as they don't have links. 
Thus, we introduce an edge generator to model the existence of edges among nodes. 
As GNNs need to learn how to extract and propagate features simultaneously, this edge generator can provide relation information for those synthesized samples, and hence facilitate the training of GNN-based classifier. This generator is trained on real nodes and existing edges, and is used to predict neighbor information for those synthetic nodes. These new nodes and edges will be added to the initial adjacency matrix $\mathbf{A}$, and serve as input the the GNN-based classifier.

In order to maintain model's simplicity and make the analysis easier, we adopt a vanilla design, weighted inner production, to implement this edge generator as:
\begin{equation}
\begin{aligned}
    \mathbf{E}_{v,u} &= softmax(\sigma(\mathbf{h}_v^1 \cdot \mathbf{S} \cdot \mathbf{h}_u^1)).
\end{aligned}
\end{equation}
where $\mathbf{E}_{v,u}$ refers to the predicted relation information between node $v$ and $u$, and $\mathbf{S}$ is the parameter matrix capturing the interaction between nodes. The loss function for training the edge generator is
\begin{equation}
\begin{aligned}
    \mathcal{L}_{edge} = \|\mathbf{E} - \mathbf{A}\|_F^2,
\end{aligned}
\end{equation}
where $\mathbf{E}$ refers to predicted connections between nodes in $\mathcal{V}$, i.e., no synthetic nodes. 
Since we learn an edge generator which is good at reconstructing the adjacency matrix using the node representations, it should give good link predictions for synthetic nodes. 

With the edge generator, we attempt two strategies to put the predicted edges for synthetic nodes into the augmented adjacency matrix. 
In the first strategy, this generator is optimized using only edge reconstruction, and the edges for the synthetic node $v'$ is generated by setting a threshold $\eta$:
\begin{equation}
    \tilde{\mathbf{A}}[v', u] = 
    \begin{cases}
    1, & \text{if } \mathbf{E}_{v', u} > \eta \\
    0,              & \text{otherwise}.
\end{cases}
\end{equation}
where $\tilde{\mathbf{A}}$ is the adjacency matrix after over-sampling, by inserting new nodes and edges into $\mathbf{A}$, and will be sent to the classifier. 

In the second strategy, for synthetic node $v'$, we use soft edges instead of binary ones:
\begin{equation}
    \tilde{\mathbf{A}}[v', u] = \mathbf{E}_{v', u},
\end{equation}
In this case, gradient on $\tilde{\mathbf{A}}$ can be propagated from the classifier, and hence the generator can be optimized using both edge prediction loss and node classification loss, which will be introduced later. Both two strategies are implemented, and their performance are compared in the experiment part.

\subsection{GNN Classifier}
Let $\tilde{\mathbf{H}}^1$ be the augmented node representation set by concatenating $\mathbf{H}^1$ (embedding of real nodes) with the embedding of the synthetics nodes, and $\tilde{\mathcal{V}}_L$ be the augmented labeled set by incorporating the synthetic nodes into $\mathcal{V}_L$. Now we have an augmented graph 
$ {\tilde{\mathcal{G}}}=\{\tilde{\mathbf{A}},\tilde{\mathbf{H}}\}$ with labeled node set $\tilde{\mathcal{V}}_L$. The data size of different classes in $\tilde{\mathcal{G}}$ becomes balanced, and an unbiased GNN classifier would be able to be trained on that.
Specifically, we adopt another GraphSage block, appended by a linear layer for node classificaiton on $\tilde{G}$ as:
\begin{equation}
\mathbf{h}_{v}^{2}  = \sigma(\mathbf{W}^2 \cdot CONCAT(\mathbf{h}_{v}^{1}, \tilde{\mathbf{H}}^{1} \cdot \tilde{\mathbf{A}}[:,v])),
\end{equation}
\begin{equation}
\mathbf{P}_{v}  = softmax(\sigma(\mathbf{W}^c \cdot CONCAT(\mathbf{h}_{v}^{2}, \mathbf{H}^{2} \cdot \tilde{\mathbf{A}}[:,v]))),
\end{equation}
where $\mathbf{H}^2$ represents node representation matrix of the 2nd GraphSage block, and $\mathbf{W}$ refers to the weight parameters. $\mathbf{P}_{v}$ is the probability distribution on class labels for node $v$. The classifier module is optimized using cross-entropy loss as:
\begin{equation}
\mathcal{L}_{node}  = \sum_{u \in \tilde{\mathcal{V}}_L} \sum_{c} ( \mathbf{1}(Y_{u}==c) \cdot log(\mathbf{P}_{v}[c]).
\end{equation}
And during testing, the predicted class for node $v$, $\mathbf{Y}_{v}^{'}$ will be set as the class with highest probability,
\begin{equation}
\mathbf{Y}_v^{'} = \argmax_{c}\mathbf{P}_{v,c}
\end{equation}

\subsection{Optimization Objective}
Putting the feature extractor, synthetic node generator, edge generator and GNN classifier together, previous parts together, the final objective function of {\method} can be written as:
\begin{equation}
\begin{aligned}
  \min_{\theta, \phi, \varphi} & \mathcal{L}_{node} + \lambda \cdot \mathcal{L}_{edge},
\end{aligned}
\end{equation}
wherein $\theta, \phi, \varphi$ are the parameters for feature extractor, edge generator, and node classifier respectively. As the model's performance is dependent on the quality of embedding space and generated edges, to make training phrase more stable, we also tried pre-training feature extractor and edge generator using $\mathcal{L}_{edge}$. 

The design of {\method} has several advantages: (i) it is easy to implement synthetic minority over-sampling process. Through uniting interpolated node embedding and predicted edges, new samples can be generated; (ii) the feature extractor is optimized using training signal from both node classification task and edge prediction task. Therefore, rich intra-class and inter-class relation information would be encoded in the embedding space, making the interpolation more robust; and (iii) it is a general framework. It can cope with different structure choices for each component, and different regularization terms can be enforced to provide prior knowledge. 

\subsection{Training Algorithm}
The full pipeline of running our framework can be summarized in Algorithm 1. Inside each optimization step, we first obtain node representations using the feature extractor in line $6$. Then, from line $7$ to line $11$, we perform over-sampling in the embedding space to make node classes balanced. After predicting edges for generated new samples in line $12$, the following node classifier can be trained on top of that over-sampled graph. The full framework is trained altogether with edge prediction loss and node classification loss, as shown in line $13$. 

\begin{algorithm}
  \caption{Full Training Algorithm}
  \label{alg:Framwork}
  \begin{algorithmic}[1] 
  \REQUIRE 
    $\mathcal{G} = \{\mathcal{V}, \mathbf{A}, \mathbf{F}, \mathbf{Y}\} $
  \ENSURE $\text{Predicted node class } \mathbf{Y}^{'}$
    \STATE Randomly initialize the feature extractor, edge generator and node classifier;
    \IF{Require pre-train}
    \STATE Fix other parts, train the feature extractor and edge generator module until convergence, based on loss $\mathcal{L}_{edge}$;
    \ENDIF
    \WHILE {Not Converged}
    \STATE Input $\mathcal{G}$ to feature extractor, obtaining $\mathbf{H}^{1}$; 
    \FOR{class c in minority classes}
    \FOR{i in $size(c) \cdot \text{over-sampling scale}$}
    \STATE Generate a new sample in class c, Following Equation (3) and (4);
    \ENDFOR
    \ENDFOR
    \STATE Generate $\mathbf{A}^{'}$ using edge generator, basing on Equation (7) or (8);
    \STATE Update the model using $\mathcal{L}_{node} + \lambda \cdot \mathcal{L}_{edge}$; 
    
    \ENDWHILE
    \RETURN Trained feature extractor, edge predictor, and node classifier module.
  \end{algorithmic}
\end{algorithm}



\section{Experiments} \label{sec:experiments}
In this section, we conduct experiments to evaluate the benefits of {\method} for the node classification task when classes are imbalanced. Both artificial and genuine imbalanced datasets are used, and different configurations are adopted to test its generalization ability. Particularly, we want to answer the following questions:
\begin{itemize}
    \item How effective is {\method} in imbalanced node classification task?
    \item How different choices of over-sampling scales would affect the performance of {\method}?
    \item Can {\method} generalize well to different imbalance ratios, or different base model structures?
\end{itemize}
We begin by introducing the experimental settings, including datasets, baselines, and evaluation metrics. We then conduct experiments to answer these questions.

\subsection{Experimental Settings}
\subsubsection{Datasets}
We conduct experiments on two widely used publicly available datasets for node classification, Cora~\cite{Sen2008CollectiveCI} and BlogCatalog~\cite{tang2009relational}, and one fake account detection dataset, Twitter~\cite{mohammadrezaei2018identifying}. The details of these three datasets are given as follows:
\begin{itemize}
    \item \textbf{Cora}: Cora is a citation network dataset for transductive learning setting. It contains one single large graph with $2,708$ papers from $7$ areas. Each node has a $1433$-dim attribution vector, and a total number of $5,429$ citation links exist in that graph. In this dataset, class distributions are relatively balanced, so we use an imitative imbalanced setting: three random classes are selected as minority classes and down-sampled. All majority classes have a training set of $20$ nodes. For each minority class, the number is $20 \times imbalance\_ratio$. We vary $imbalance\_ratio$ to analyze the performance of {\method} under various imbalanced scenarios.
    \item \textbf{BlogCatalog}: This is a social network dataset crawled from BlogCatalog\footnote{http://www.blogcatalog.com}, with $10,312$ bloggers from $38$ classes and $333,983$ friendship edges. The dataset doesn't contain node attributes. Following ~\cite{Perozzi2014DeepWalkOL}, we attribute each node with a $64$-dim embedding vector obtained from Deepwalk. Classes in this dataset follow a genuine imbalanced distribution, with $14$ classes smaller than $100$, and $8$ classes larger than $500$. For this dataset, we use $25\%$ samples of each class for training and 25\% for validation, the remaining $50\%$ for testing.
    \item \textbf{Twitter}: This dataset is crawled by ~\cite{mohammadrezaei2018identifying} with a dedicated API crawler from Twitter\footnote{https://twitter.com} on bot infestation problem. It has $5,384,160$ users in total. Among them, $63,167$ users are bots. In this work, we split a connected sub-graph from it containing $61,122$ genuine users and $2,045$ robots. Node embedding is obtained through Deepwalk, appended with node degrees. This dataset is used for binary classification, and the imbalance ratio is roughly $1:30$. We randomly select $25\%$ of total samples for training, 25\% for validation, and the remaining $50\%$ for testing.
\end{itemize}

\subsubsection{Baselines}
We compare {\method} with representative and state-of-the-art approaches for handling imbalanced class distribution problem, which includes:
\begin{itemize}
    \item Over-sampling: A classical approach for imbalanced learning problem, by repeating samples from minority classes. We implement it in the raw input space, by duplicating $n_s$ minority nodes along their edges. In each training iteration, $\mathcal{V}$ is over-sampled to contain $n+n_s$ nodes, and $\mathbf{A} \in \mathbb{R}^{(n+n_s) \times (n+n_s)}$.
    \item Re-weight~\cite{Yuan2012SamplingR}: This is a cost-sensitive approach which gives class-specific loss weight. In particular, it assigns higher loss weights to samples from minority so as to alleviate the issue of majority classes dominating the loss function.
    \item SMOTE~\cite{chawla2002smote}: Synthetic minority oversampling techniques generate synthetic minority samples by interpolating a minority samples and its nearest neighbors of the same class. For newly generated nodes, its edges are set to be the same as the target node.  
    \item Embed-SMOTE~\cite{ando2017deep}: An extension of SMOTE for deep learning scenario, which perform over-sampling in the intermediate embedding layer instead of the input domain. We set it as the output of last GNN layer, so that there is no need to generate edges.
\end{itemize}

Basing on the strategy for training edge generator and setting edges, four implementations of {\method} are tested:
\begin{itemize}
    \item ${\method}_T$: The edge generator is trained using loss from only edge prediction task. The predicted edges are set to binary values with a threshold before sending to GNN-based classifier;
    \item ${\method}_O$: Predicted edges are set as continuous so that gradient can be calculated and propagated from GNN-based classifier. The edge generator is trained along with other components with training signals from both edge generation task and node classification task;
    \item ${\method}_{preT}$: An extension of ${\method}_T$, in which the feature extractor and edge generator are pre-trained on the edge prediction task, before fine-tuning on Equation.13. During fine-tuning, edge generator is optimized using only $\mathcal{L}_{edges}$;
    \item ${\method}_{preO}$: An extension of ${\method}_O$, in which a pre-training process is also conducted before fine-tuning, same as ${\method}_{preT}$.
\end{itemize}

In the experiments, all these methods are implemented and tested on the same GNN-based network for a fair comparison.

\begin{table*}[t]
  \setlength{\tabcolsep}{4.5pt}
  \normalsize
  \small
  
  \caption{Comparison of different approaches for imbalanced node classification.}\label{tab:result}
  \vskip -1em
  \begin{tabular}{p{2.1cm} | p{1.40cm}  p{1.40cm}  p{1.40cm} | p{1.40cm}  p{1.40cm}  p{1.40cm} | p{1.40cm}  p{1.40cm}  p{1.40cm} }

    \hline
     &  \multicolumn{3}{|c|}{Cora} &  \multicolumn{3}{|c|}{BlogCatalog} &  \multicolumn{3}{|c}{Twitter} \\
    \hline
    Methods & ACC & AUC-ROC & F Score & ACC & AUC-ROC & F Score & ACC & AUC-ROC & F Score \\
    \hline
    Origin & $0.681\pm0.001$ & $0.914\pm0.002$ & $0.684\pm0.003$ & $0.210\pm0.004$ & $0.586\pm0.002$ & $0.074\pm0.002$ & $0.967\pm0.004$ & $0.577\pm0.003$ & $0.494\pm0.001$  \\
    over-sampling & $0.692\pm0.009$ & $0.918\pm0.005$ & $0.666\pm0.008$ & $0.203\pm0.004$ & $0.599\pm0.003$ & $0.077\pm0.001$ & $0.913\pm0.006$ & $0.601\pm0.011$ & $0.513\pm0.003$  \\
    Re-weight & $0.697\pm0.008$ & $0.928\pm0.005$ & $0.684\pm0.004$ & $0.206\pm0.005$ & $0.587\pm0.003$ & $0.075\pm0.003$ & $0.915\pm0.005$ & $0.603\pm0.004$ & $0.515\pm0.002$  \\
    SMOTE & $0.696\pm0.011$ & $0.920\pm0.008$ & $0.673\pm0.003$ & $0.205\pm0.004$ & $0.595\pm0.003$ & $0.077\pm0.001$ & $0.914\pm0.005$ & $0.604\pm0.007$ & $0.514\pm0.002$  \\
    Embed-SMOTE & $0.683\pm0.007$ & $0.913\pm0.002$ & $0.673\pm0.002$ & $0.205\pm0.003$ & $0.588\pm0.002$ & $0.076\pm0.001$ & $0.943\pm0.004$ & $0.606\pm0.005$ & $0.514\pm0.002$  \\
    \hline
    ${\method}_T$ & $0.713\pm0.008$ & $0.929\pm0.006$ & $0.720\pm0.002$ & $0.206\pm0.005$ & $0.602\pm0.004$ & $0.083\pm0.003$ & $0.929\pm0.005$ & $0.622\pm0.003$ & $0.519\pm0.001$ \\
    ${\method}_O$ & $0.709\pm0.010$ & $0.927\pm0.011$ & $0.712\pm0.003$ & $0.215\pm0.010$ & $0.591\pm0.012$ & $0.080\pm0.005$ & $0.905\pm0.008$ & $0.616\pm0.006$ & $0.515\pm0.003$ \\
    ${\method}_{preT}$ & $0.727\pm0.003$ & $0.931\pm0.002$ & $0.726\pm0.001$  & $\mathbf{0.249}\pm0.002$ & $\mathbf{0.641}\pm0.001$ & $\mathbf{0.126}\pm0.001$ & $0.937\pm0.003$ & $\mathbf{0.639}\pm0.002$ & $0.531\pm0.001$ \\
    ${\method}_{preO}$ & $\mathbf{0.736}\pm0.001$ & $\mathbf{0.934}\pm0.002$ & $\mathbf{0.727}\pm0.001$ & $0.243\pm0.002$ & $\mathbf{0.641}\pm0.002$ & $0.123\pm0.001$ & $\mathbf{0.941}\pm0.002$ & $0.636\pm0.001$ & $\mathbf{0.532}\pm0.001$ \\
    \hline
  \end{tabular}
\end{table*}

\subsubsection{Evaluation Metrics}
Following existing works in evaluating imbalanced classification~\cite{rout2018handling,johnson2019survey}, we adopt three criteria: classification accuracy(ACC), mean AUC-ROC score~\cite{bradley1997use}, and mean F-measure. ACC is computed on all testing examples at once, therefore may underweight those under-represented classes. AUC-ROC score illustrates the probability that the corrected class is ranked higher than other classes, and F-measure gives the harmonic mean of precision and recall for each class. Both AUC-ROC score and F-measure are calculated separately for each class and then non-weighted average over them, therefore can better reflect the performance on minority classes. 

\subsubsection{Configurations}
All experiments are conducted on a $64$-bit machine with Nvidia GPU (Tesla V100, 1246MHz , 16 GB memory), and ADAM optimization algorithm is used to train all the models.

For all methods, the learning rate is initialized to $0.001$, with weight decay being $5e-4$. $\lambda$ is set as $1e-6$, since we did not normalize $\mathcal{L}_{edge}$ and it is much larger than $\mathcal{L}_{node}$. On Cora dataset, imbalance\_ratio is set to $0.5$ and over-sampling scale is set as $2.0$ if not specified otherwise. For BlogCatalog and Twitter dataset, imbalance\_ratio is not involved, and over-sampling scale is set class-wise: $\frac{n}{m \cdot |\mathcal{C}_i|}$ for minority class $i$, to make the class size balanced. Besides, all models are trained until converging, with the maximum training epoch being $5000$.

\subsection{Imbalanced Classification Performance}
To answer the first question, we compare the imbalanced node classification performance of {\method} with the baselines on aforementioned three datasets. Each experiment is conducted 3 times to alleviate the randomness. The average results with standard deviation are reported in Table~\ref{tab:result}. From the table, we can make following observations:
\begin{itemize}
    \item All four variants of {\method} showed significant improvements on imbalanced node classification task, compared to the ``Origin'' setting, in which no special algorithm is adopted. They also outperform almost all baselines in all datasets, on all evaluation metrics. These results validate the effectiveness of proposed framework.
    \item The improvements brought by {\method} are much larger than directly applying previous over-sampling algorithms. For example, compared with Over-sampling ${\method}_T$ shows an improvement of $0.011, 0.003, 0.021$ in AUC-ROC score, and an improvement of $0.016, 0.014,  0.016$ in AUC-ROC score compared with Embed-SMOTE. This result validates the advantages of {\method} over previous algorithms, in constructing an embedding space for interpolation and provide relation information.
    \item Among different variants of {\method}, pre-trained implementations show much stronger performance than not pre-trained ones. This result implies the importance of a better embedding space in which the similarities among samples are well encoded.
\end{itemize}

To summarize, these results prove the advantages of introducing over-sampling algorithm for imbalanced node classification task. They also validate that {\method} can generate more realistic samples and the importance of providing relation information.

\begin{figure}[t!]
  \centering
    \includegraphics[width=0.45\textwidth]{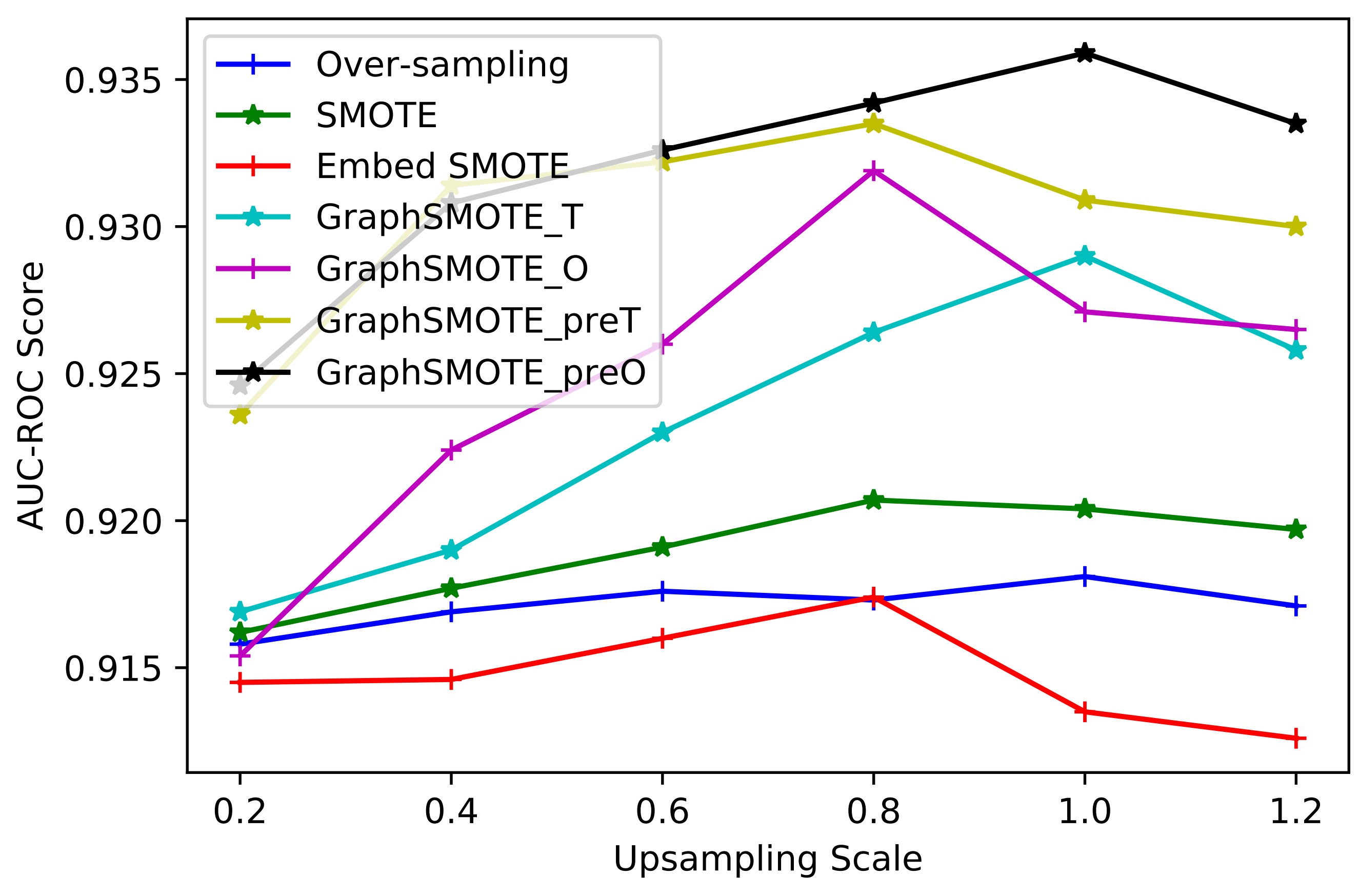}
    \vskip -1em
    \caption{Affects of over-sampling scale.} \label{fig:UpScale}
  \setlength{\abovecaptionskip}{0cm}
\end{figure}

\subsection{Influence of Over-sampling Scale}
In this subsection, we analyze the performance change of different algorithms w.r.t different over-sampling scales, in the pursuit of answering the second question. To conduct experiments in a constrained setting, we use Cora dataset and fix imbalance ratio as $0.5$. Over-sampling scale is varied as $\{0.2, 0.4, 0.6, 0.8, 1.0, 1.2\}$. Every experiment is conducted 3 times and the average results are presented in Figure~\ref{fig:UpScale}. From the figure, we make the following observations:
\begin{itemize}
    \item When over-sampling scale is smaller than $0.8$, generating more samples for minority classes, i.e., making the classes more balanced, would help the classifier to achieve better performance, which is as expected because these synthetic nodes not only balance the datasets but also introduce new supervision for training a better GNN classifier.
    \item When the over-sampling scale becomes larger, keeping increasing it may result in opposite effects. It can be observed that the performance remains similar, or degrade a little when changing over-sampling scale from $1.0$ to $1.2$. This is because when too many synthetic nodes are generated, some of these synthetic nodes contain similar/redundant information which cannot further help learn a better GNN.
    \item Based on these observations, generally setting the over-sampling scale set a value that can make the class balanced is a good choice, which is consistent with existing work for synthetic minority oversampling~\cite{buda2018systematic}.
\end{itemize}

\subsection{Influence of Imbalance Ratio}
In this subsection, we analyze the performance of different algorithms with respect to different imbalance ratios, to evaluate their robustness. Experiment is also conducted in a well-constrained setting on Cora, by fixing over-sampling scale to $1.0$, and varying imbalance ratio as $\{0.1,0.2,0.4,0.6\}$. Each experiments are conducted 3 times and the average results are shown in Table~\ref{tab:ImRatio}. From the table, we make the following observations:
\begin{itemize}
    \item The proposed framework {\method} generalizes well to different imbalance ratios. It achieves the best performance across all the settings, which shows the effectiveness of the proposed framework under various scenarios.
    \item The improvement of {\method} is more significant when the imbalance extent is more extreme. For example, when imbalance ratio is $0.1$, ${\method}_{preO}$ outperforms Re-weight by $0.0326$, and the gap reduces to $0.0060$ when the imbalance ratio become $0.6$. This is because when the datasets is not that imbalanced, minority oversampling is not that important, which makes the improvement of proposed algorithm over others not that significant.
    \item Pre-training is important when the imbalance ratio is extreme. When imbalance ratio is $0.1$, ${\method}_{preO}$ shows an improvement of $0.0268$ over ${\method}_{preO}$, and the gap reduces to $0.0055$ when the imbalance ratio changes to $0.6$.
\end{itemize}

\begin{table}[t!]
  \setlength{\tabcolsep}{4.5pt}
  \normalsize
  \caption{Node classification performance in terms of AUC on Cora under various imbalance ratios.} \label{tab:ImRatio}
  \vskip -1em
  \begin{tabular}{c || c | c | c | c  }
    \hline
     &  \multicolumn{4}{|c}{Imbalance Ratio} \\
    \hline
    Methods & $0.1$ & $0.2$ & $0.4$ & $0.6$ \\
    \hline
    Origin & $0.8681$ & $0.8998$ & $0.9139$ & $0.9146$ \\
    over-sampling & $0.8707$ & $0.9039$ & $0.9137$ & $0.9215$ \\
    Re-weight & $0.8791$ & $0.8881$ & $0.9257$ & $0.9306$ \\
    SMOTE & $0.8742$ & $0.9027$ & $0.9161$ & $0.9237$ \\
    Embed-SMOTE & $0.8651$ & $0.8967$ & $0.9188$ & $0.9212$ \\
    \hline
    ${\method}_T$ & $0.8824$ & $\mathbf{0.9162}$ & $0.9262$ & $0.9309$ \\
    ${\method}_O$ & $0.8849$ & $0.9061$ & $0.9216$ & $0.9311$ \\
    ${\method}_{preT}$ & $\mathbf{0.9167}$ & $0.9130$ & $0.9303$ & $0.9317$ \\
    ${\method}_{preO}$ & $0.9117$ & $0.9116$  & $\mathbf{0.9389}$ & $\mathbf{0.9366}$ \\
    \hline
  \end{tabular}
\end{table}

\subsection{Influence of Base Model}
In this subsection, we test generalization ability of the proposed algorithm by applying it to another widely-used graph neural network: GCN. Comparison between it and baselines is presented in Table~\ref{tab:GCN}. All methods are implemented on the same network. Experiments are performed on Cora, with imbalance ratio set as $0.5$ and over-sampling scale as $2.0$. Experiments are run three times, with both averaged results and standard deviation reported. From the result, it can be observed that: 
\begin{itemize}
    \item Generally, {\method} adapt well to GCN-based model. Four variants of it all work well and achieve the best performance, as shown in Table~\ref{tab:GCN}.
    \item Compared with using GraphSage as base model, a main difference is that pre-training seems to be less necessary in this case. We think it may be caused by the fact that GCN is less powerful than GraphSage in representation ability. GraphSage is more flexible and can model more complex relation information, and hence is more difficult to train. Therefore, it can benefit more from obtaining a well-trained embedding space in advance.
\end{itemize}

\begin{table}[h!]
  \setlength{\tabcolsep}{4.5pt}
  \normalsize
  \caption{Evaluation of different algorithm's performance when changed to GCN as base model. } \label{tab:GCN}
  \vskip -1em
  \begin{tabular}{c || c | c | c   }
    \hline
     &  \multicolumn{3}{|c}{Cora} \\
    \hline
    Methods & ACC & AUC-ROC & F Score  \\
    \hline
    Origin & $0.685\pm0.002$ & $0.907\pm0.003$ & $0.663\pm0.001$ \\
    over-sampling & $0.682\pm0.005$ & $0.907\pm0.003$ & $0.665\pm0.003$ \\
    Re-weight & $0.684\pm0.005$ & $0.913\pm0.004$ & $0.672\pm0.002$ \\
    SMOTE & $0.684\pm0.006$ & $0.910\pm0.005$ & $0.665\pm0.003$ \\
    Embed-SMOTE & $0.691\pm0.002$ & $0.910\pm0.003$ & $0.667\pm0.002$  \\
    \hline
    ${\method}_T$ & $0.695\pm0.005$ & $\mathbf{0.920\pm0.003}$ & $0.690\pm0.002$  \\
    ${\method}_O$ & $0.693\pm0.005$ & $0.920\pm0.005$ & $\mathbf{0.707}\pm0.003$  \\
    ${\method}_{preT}$ & $0.688\pm0.001$ & $0.919\pm0.002$ & $0.682\pm0.001$ \\
    ${\method}_{preO}$ & $\mathbf{0.699}\pm0.002$ & $0.914\pm0.002$  & $0.702\pm0.001$  \\
    \hline
  \end{tabular}
\end{table}

\subsection{Parameter Sensitivity Analysis}
\begin{figure}[t!]
  \centering
  \subfigure[AUC-ROC Score]{
		\label{fig:auc-lambda}
		\includegraphics[width=0.23\textwidth]{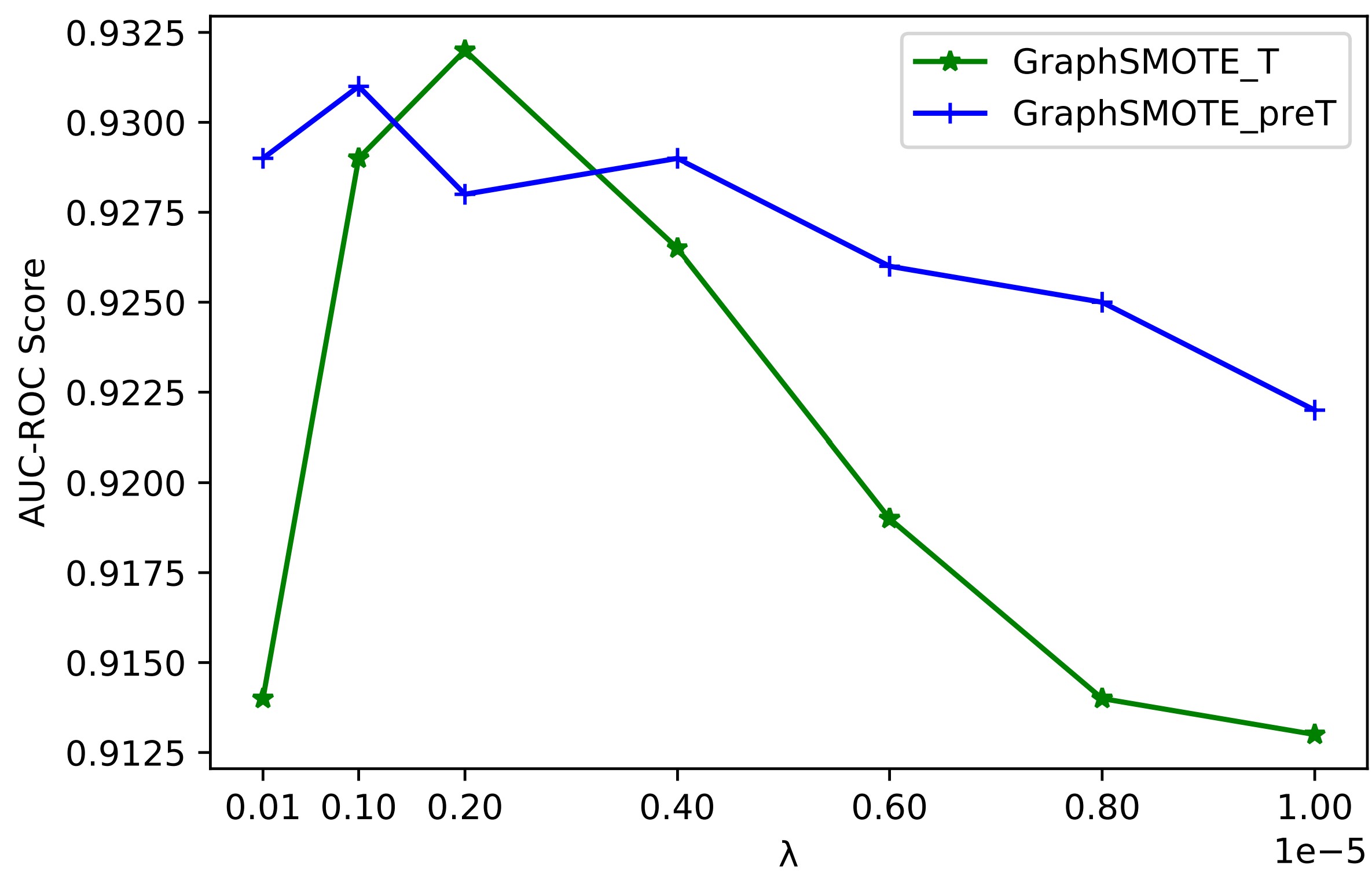}}
    \subfigure[F Measurement]{
		\label{fig:f-lambda}
		\includegraphics[width=0.23\textwidth]{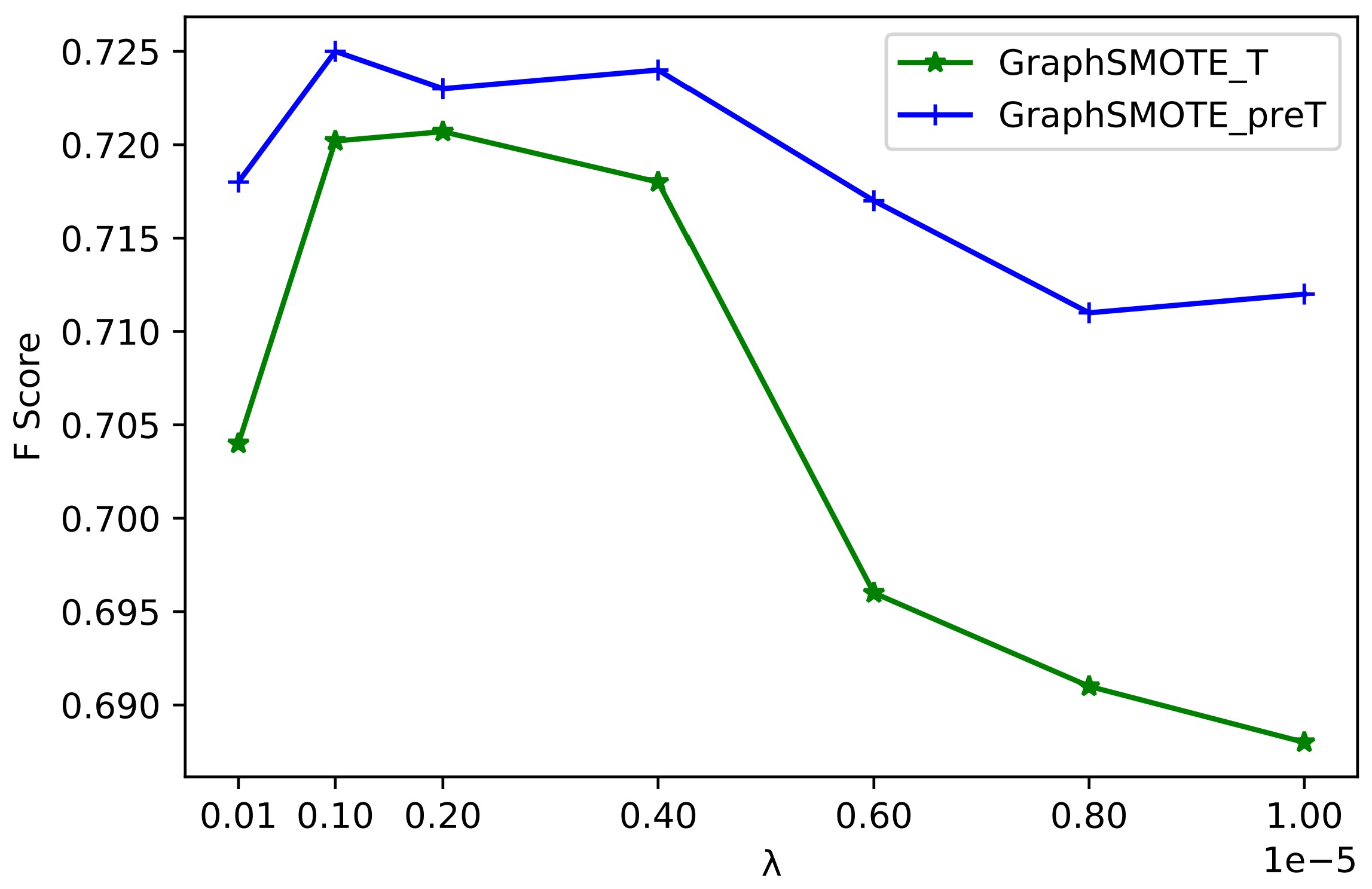}}
		
    \vskip -1em
    \caption{Affects of hyper-parameter $\lambda$.} \label{fig:lambda}
  \setlength{\abovecaptionskip}{0cm}
\end{figure}

In this part, the hyper-parameter $\lambda$ is varied to test {\method}'s sensitivity towards it. To keep simplicity, we adopt ${\method}_T$ and ${\method}_{preT}$ as base model, and set $\lambda$ to be in $\{1e-7,1e-6,2e-6,4e-6,6e-6,8e-6,1e-5\}$. Each experiment is conducted on Cora with imbalance ratio $0.5$ and over-sampling scale $2.0$. The results weare shown in Figure~\ref{fig:lambda}. From the figure, we can observe that:
\begin{itemize}
    \item Generally, as $\lambda$ increases, the performance first increase then decrease. The performance would drop significantly if $\lambda$ is too large. Generally, a smaller $\lambda$ between $1e-6$ and $4e-6$ works better. The reason could be the difference in scale of two losses.
    \item Pre-training makes {\method} more stable w.r.t $\lambda$.
\end{itemize}

\section{Conclusion and Future Work} \label{sec:conclusion}
Class imbalance problem of nodes in graphs widely exists in real-world tasks, like fake user detection, web page classification, malicious machine detection, etc. This problem can significantly influence classifier's performance on those minority classes, but is left unconsidered in previous works. Thus, in this work, we investigate this imbalanced node classification task. Specifically, we propose a novel framework {\method}, which extends previous over-sampling algorithms for i.i.d data to this graph setting. Concretely, {\method} constructs an intermediate embedding space with a feature extractor, and train an edge generator and a GNN-based node classifier simultaneously on top of that. Experiments on one artificial dataset and two real-world datasets demonstrated its effectiveness, outperforming all other baselines with a large margin. Ablation studies are performed to understand {\method} performs under various scenarios. Parameter sensitivity analysis is also conducted to understand the sensitivity of {\method} on the hyperparameters.

There are several interesting directions need further investigation. First, besides node classification, other tasks like edge type prediction or node representation learning may also suffer from under-representation of nodes in minority classes. And sometimes, node class might not be provided explicitly. Therefore, we will also extend {\method} for handling other types of imbalanced learning problems on graphs. Second, in this paper, we mainly conduct experiments on citation network and social media network. There are many other real-world applications which can be treated as imbalanced node classification problems. Therefore, we would like to extend our framework for more application domains such as document analysis in the websites.

\section{Acknowledgement}
This project was partially supported by NSF projects IIS-1707548, CBET-1638320, IIS-1909702, IIS1955851, and the Global Research Outreach program of Samsung Advanced Institute of Technology under grant \#225003.

\bibliographystyle{ACM-Reference-Format}
\bibliography{acmart}

\end{document}